\documentclass[a4paper]{article}

\usepackage{INTERSPEECH2016}

\usepackage{graphicx}
\usepackage{graphics}
\usepackage{amssymb,amsmath,bm}
\usepackage{textcomp}
\usepackage{booktabs}
\usepackage{multirow}
\usepackage{cite}

\usepackage{color}
\definecolor{cardinal}{cmyk}{0,1,0.63,0.29}

\ninept

\newcommand{\eg}{\textit{e.g.,\ }}
\newcommand{\etc}{\textit{etc.\@}}

\newcommand{\ie}{\textit{i.e.,\ }}
\newcommand{\etal}{\textit{et al.\@}}

\newcommand{\Pg}[1]{\noindent\emph{\textbf{\color{cardinal}#1}}}%paragraph

\usepackage[%
breaklinks=true,
colorlinks=false, % For Printed version make false
citecolor=cardinal,
urlcolor=cardinal,
linkcolor=cardinal,
linktocpage,
draft=false,
pageanchor=true,
hyperfigures=true,
plainpages=false,
pagebackref=false
]{hyperref}

\title{Sparsely Connected and Disjointly Trained Deep Neural Networks for Low Resource Behavioral Annotation: Acoustic Classification in Couples' Therapy}
\makeatletter
\def\name#1{\gdef\@name{#1\\}}
\makeatother \name{{\em Haoqi Li$^1$, Brian Baucom$^2$, Panayiotis Georgiou$^1$}}

\address{$^1$University of Southern California, Los Angeles, CA, USA \\
  $^2$The University of Utah, Department of Psychology, UT, USA \\
  {\small \tt haoqili@usc.edu, brian.baucom@utah.edu, georgiou@sipi.usc.edu}
}

\begin{document}
%\nbib
  \maketitle
  \begin{abstract}
    % length is limited to 200 words
    Observational studies are based on accurate assessment of human state. 
    A behavior recognition system that models interlocutors' state in real-time can significantly aid the mental health domain.
    However, behavior recognition from speech remains a challenging task since it is difficult to find generalizable and representative features because of noisy and high-dimensional data, especially when data is limited and annotated coarsely and subjectively.
    Deep Neural Networks (DNN) have shown promise in a wide range of machine learning tasks, but for Behavioral Signal Processing (BSP) tasks their application has been constrained due to limited quantity of data. 
    
    We propose a Sparsely-Connected and Disjointly-Trained DNN (SD-DNN) framework to deal with limited data. 
    First, we break the acoustic feature set into subsets and train multiple distinct classifiers. 
    Then, the hidden layers of these classifiers become parts of a deeper network that integrates all feature streams. 
    The overall system allows for full connectivity while limiting the number of parameters trained at any time and allows convergence possible with even limited data. 
    We present results on multiple behavior codes in the couples' therapy domain and demonstrate the benefits in behavior classification accuracy. We also show the viability of this system towards live behavior annotations.
  \end{abstract}
  
\section{Introduction}
Observational practice, such as in the field of psychology, relies heavily on analysis of human behaviors based on observable interaction cues. 
In Couples' Therapy, one fundamental task is to observe, evaluate and identify domain-specific behaviors during couples’ interactions. 
Based on behavioral analyses, psychologists can provide effective and specific treatment. 
	
Rating behaviors by human annotators is a costly and time consuming process. 
Great advances have been made during last decade on assessing human state through technical way. 
For example, speech emotion recognition works~\cite{el2011survey, Vogt2008, Schuller20111062} have shown effectiveness of extracting emotional content from human speech signals. 
In addition, Deep Neural Networks (DNN) have been employed for many related speech tasks\cite{6296526, 5947651, Schuller2015}. 
Han \etal \cite{han2014speech} and Le \etal \cite{6707732} both utilized DNN to extract high level representative features to improve emotion classification accuracy.

Human emotions can change quickly and frequently in a short time period, thus emotion recognition mainly focuses on very short speech segments (\eg less than 2s). 
Affect recognition models basic emotions and is not domain-specific. 
For mental health applications, though, experts are more interested in very specific and complex behaviors exhibited over longer time scales.
Over the last few years Behavioral Signal Processing (BSP) \cite{6457406,Georgiou:2011:BSP} has examined the analysis of such complex, domain specific behaviors.
Based on machine learning techniques, BSP employed lexical \cite{Georgiou2011}, acoustic \cite{Black20131}, and visual \cite{7106538,6607640} information to analyze and model multimodal human behaviors. 
For instance, in couples' therapy domain, Black \etal \cite{Black20131} built an automatic human behavioral coding system for couples’ interaction by using acoustic features. 
In \cite{7178339,xia2015dynamic} the authors employed a top layer HMM to take dynamic behavior state transitions into consideration and thus achieved higher accuracy on session-level behavioral classification.

Despite these efforts, behavior estimation is still a complex task. 
Session level models combine information at different timescales to estimate a session level rating.
In doing so, they ignore non-linear information integration models which are often employed by human raters, such as recency and primacy models. % see https://en.wikipedia.org/wiki/Serial_position_effect#Primacy_effect
%For one thing, how human interpret and integrate behavior information is still not well understood since the process is not a simple linear system; %(Maybe ask Brian to provide some references)
Further, and one of the biggest challenges, is that representative samples of behavior are extremely limited due to privacy constraints, cost of annotation, subjective ground truth, and coarse annotations (both attributed to cost and human contextualization of short-term information). 

Deep Neural Networks have shown promise in a wide range of machine learning tasks, especially for their ability to extract high level descriptions from raw data. 
However, in BSP, due to the limited quantity of data, DNN deployment is difficult.
% in addition?  
Because of limited data, high-dimensionality acoustic features, high signal variability, and the complication that the same acoustic signal encodes a range of additional information, training DNN systems on such data fails to converge to optimal operating conditions.

To address this problem, we propose a Sparsely-Connected and Disjointly-Trained Deep Neural Networks (SD-DNN) and demonstrate its use for behavioral recognition in Couples' Therapy. 

%% (Comments: should we make more description of our method here???)
%This paper utilized DNN as classifier to predict behavioral in four dimensions (Acceptance, Positive, Negative, Blame).  
%DNN is supposed to model complex and non-linear high-level representation of low-level acoustic features, and finally improve the accuracy of learning system. 
%However,We address this BSP data-sparsity challenge by introducing a new DNN joint co-training structure: instead of using all features at the beginning, we utilize subset of features to train base learning system separately. 
%After each sub-learning system converges, we fixed those weights to reduce the flexibility of learning system, and build fusion layers on top of multiple base layers to classify human behaviors. Results show improvement of this training structure compare with plain DNN, especially in dealing with sparsity issue of training data set.

The rest of our paper is organized as follows: 
Section 2 describes audio pre-processing steps and feature extraction methods employed in our work.
Section 3 provides a brief description of the database used in experiments.
Section 4 describes the proposed SD-DNN behavior learning system in detail, after which we design multiple experiments and discuss our results in Section 5 and 6. 
Finally, we present our conclusions in Section 7.
 
\section{Preprocessing and feature extraction}
\subsection{Audio preprocessing}
%\panos{
%THIS SECTION SHOULD BE ONLY 5-6 LINES. YOU SHOULD RELY ON/CITE MATT'S PAPER FOR DETAILS.SOMETING ALONG THE LINES OF:}
In any acoustic behavior classification task, we first need to identify contiguous regions of speech by the interlocutors. 
This requires a range of pre-processing steps: \textit{Voice Activity Detection} (VAD) to identify spoken regions, \textit{Speaker Diarization} to identify same-speaker regions. Following this, we perform the feature extraction from speech regions. 
In our work we employ the preprocessing steps described in \cite{Black20131}. In short: We employ all available interactions with a SNR above 5dB, and perform VAD and Diarization. 
Then we ignore speech segments that are shorter than 1.5 seconds. Speech segments from each session for the same speaker are then used to analyze behaviors.

\subsection{Acoustic feature extraction\label{sec:feature}}
We extract acoustic features characterizing speech prosody (pitch and intensity), spectral envelope characteristics (MFCCs, MFBs), and voice quality (jitter and shimmer). 
All these Low-Level-Descriptors (LLDs) are extracted every 10 \textit{ms} with a 25 \textit{ms} Hamming window through \textit{openSMILE}\cite{opensmile} and \textit{PRAAT}\cite{boersma2002praat}. 
We perform session level feature normalization for each of the speakers as in \cite{Black20131} to reduce the impact of recording conditions and physical characteristics of different speakers. 

Unlike \cite{Black20131} we are interested in building a fine-resolution behavioral estimation, rather than session-level classification-only system, and as such we employ features with a sliding frame\footnote{Note: arguably this could be converted into an online system if the normalization was done with a slower-varying sliding window, akin to the CMV normalization of ASR systems.}.
Within each frame, we calculate a number of functionals: Min (1st percentile), Max (99th percentile), Range (99th percentile -- 1st percentile), Mean, Median, and Standard Deviation.
%  features as shown in Table \ref{features}. 
% \begin{table}[h]
% \centering
% \begin{tabular}{ll}
% \hline
% Feature Type  & Feature Names                                                                                                                                                    \\ \hline
% Prosody       & Intensity, Pitch                                                                                                                                                 \\
% Spectral      & MFCC{[}0-14{]}, MFB{[}0-7{]}                                                                                                                                     \\
% Voice quality & Jitter, Shimmer                                                                                                                                                  \\ \hline
% functionals   & \begin{tabular}[c]{@{}l@{}}Min(1st percentile), Max(99th percentile), \\ Range(99th percentile -- 1th percentile),\\ Mean, Median, Standard Deviation\end{tabular} \\ \hline
% \end{tabular}
% \caption{Acoustic features and statistical functionals.}
% \label{features}
% \end{table}

\section{Couples' Therapy Corpus}
The database used in this paper is provided by UCLA/UW Couple Therapy Research Project~\cite{christensen2004traditional}, in which 134 couples participated in video-taped problem-solving interactions. During each discussion, a relationship-related topic (e.g.``why can't you leave my stuff alone?'') was selected. 
Each participant's behaviors was rated separately by human annotators for a set of 33 behavioral codes (e.g. ``Blame'', ``Acceptance'' \etc) based on the Couples Interaction Rating System (CIRS)~\cite{heavey2002couples} and the Social Support Interaction Rating System (SSIRS)~\cite{jones1998couples}. Every human annotator provided a subjective rating scale from 1 to 9, where 1 refers absence of the behavior and 9 indicates a strong presence. For more information about this dataset, please refer to \cite{Black20131, christensen2004traditional}.

\section{Methodology}
Human experts integrate a range of cues over a wide time interval and significant context to arrive at session-level behavior descriptors. 
For example, a therapist can observe a couple interacting for an hour and derive an assessment that one of the partners is negative while the other shows acceptance.
This, unfortunately, means that we are often left without an instantaneous ground-truth.
More often than not, this results in either building session level systems by employing all available data \eg \cite{Black20131}, averaging of local decisions towards session level ratings \cite{Georgiou2011}, or creating models of interaction as in \cite{7178339,xia2015dynamic}.

In this work, we will build a system that is able to estimate behaviors over short time frames towards implementing a live behavioral estimation framework. 
We propose a Sparsely-Connected and Disjointly-Trained Deep Neural Network (SD-DNN), that aims to tackle the data sparsity issues in behavioral analysis. 

\begin{figure}[t]
  \centering
  \includegraphics[width=\linewidth]{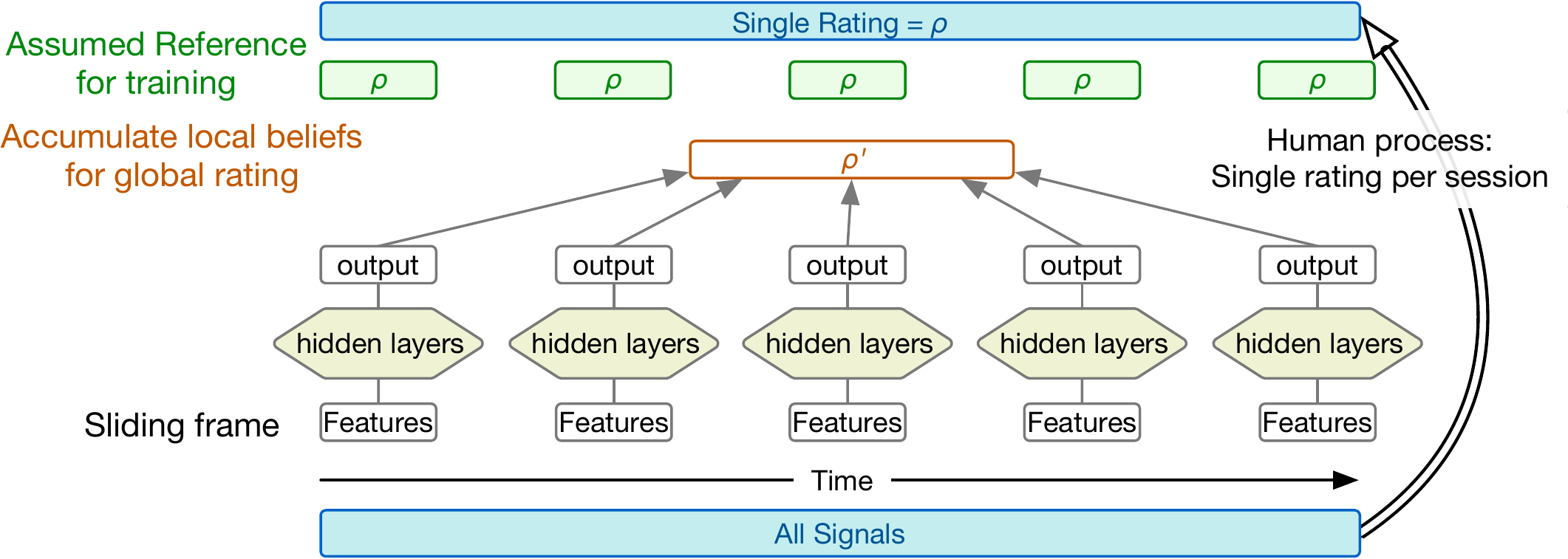}
  \caption{{\it During training, local reference is assumed to be equal to global as denoted by the green row of $\rho$. During testing the mean rating is assigned as the estimated session-level rating $\rho'$.}}
  \label{fig:OverallProcess}
\end{figure}

Due to the lack of ground truth at short time intervals, we will employ session level ratings for training and evaluation. For training, we will assume that every frame in a session shares the same rating as the session level gestalt rating as shown on Fig.~\ref{fig:OverallProcess}. For evaluation, we will use the average of the macro-coding to estimate session level coding. Finally, we will demonstrate how the system is able to track behavioral trajectories.

%Under this assumption, we segment speech audio into short time period frames to analyze presence of certain behavior at frame level, then integrate frame level decision to predict behavior at session level, and finally try to improve behavioral recognition accuracy. 
%Our proposed system,  a Sparsely-Connected and Disjointly-Trained Deep Neural Network (SD-DNN),  aims to solve the data sparsity % issue in behavioral analysis area. 
% Figure~\ref{fig:overview} shows the overview of our approach. 

% \panos{YOU HAVE TO INTRODUCE THE SYSTEMS IN SEQUENCE: 1st: fully connected DNN: Each frame gets label of session, evaluation comes through session level averaging. 2. System is broken into sd-dnn...}

% After feature extraction, we break the acoustic feature set in to subsets and train an individual classifier for each subset independently. 
% Then, using hidden layers of these classifiers and train a parent layer to integrate information from distinct classifiers.
% Comments: not sure if this figure is necessary
%\begin{figure}[t]
%  \centering
%  \includegraphics[width=\linewidth]{cut_overview.pdf}
%  \caption{{\it SD-DNN behavioral recognition algorithm overview}}
%  \label{fig:overview}
%\end{figure}

%To show the effectiveness of our training framework, we first describe difficulties of using plain DNN training method based on real experiment experience, and then describe our method in details.

\subsection{DNN training\label{sec:dnn-training}}
Employing the usual way of training a DNN system requires significant amounts of data. In our analysis, and with a feature size of 168, this approach always lead to failure during training: DNN training immediately identifies a local minimum even for small neural networks; while the objective function decreases on the training set, it does not on the development set. 
Behavioral recognition results during testing are mostly unchanging, and hence uninformative in providing behavioral trajectories. Likely the system converges to different minima relating to other dimensions, such as for instance speaker characteristics.

To minimize overfitting we can add a dropout layer\cite{dropout} at the input. This feature reduction avoids overfitting to a certain degree, however we still do not obtain the gains we expected from employing a DNN framework.

\subsection{Reduced feature dimensionality DNN}
\begin{figure}[t]
  \centering
  \includegraphics[width=\linewidth]{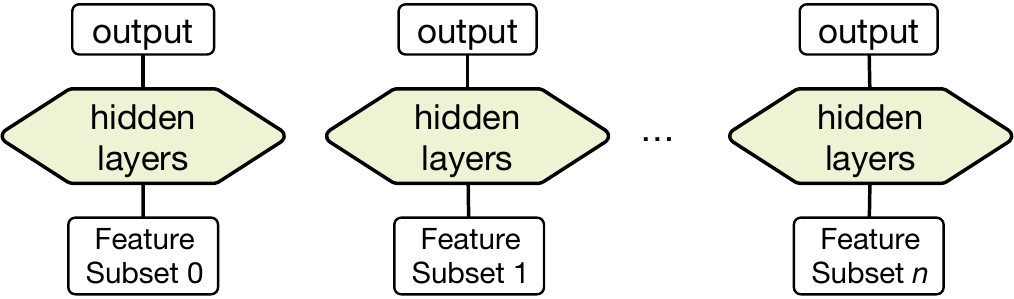}
  \caption{{\it Basic behavior recognition system based on sub feature set}}
  \label{fig:basic}
\end{figure}

One way to avoid overfitting issues is to use a reduced dimensionality input feature set. We can do that through selecting a subset of features and training DNN on those, which means we use these sub-feature-sets to train multiple behavior recognition systems. 
For each of these systems, the feature dimension is reduced by a significant factor compared to the full feature set, thus number of parameters in the resulting DNN is also decreased. Using same amount of training data, we can obtain a robustly trained DNN.
The process of this stage is shown in Figure~\ref{fig:basic}. 

As we expect, this does not perform above baseline systems either since we do not employ all informative features in to consideration.
Subsequent output fusion is also challenging and does not improve performance.

\subsection{Sparsely-Connected and Disjointly-Trained DNN\label{sec:spars-conn-disj}}
\begin{figure}[t]
  \centering
  \vspace{-0.12cm}
  \includegraphics[width=\linewidth]{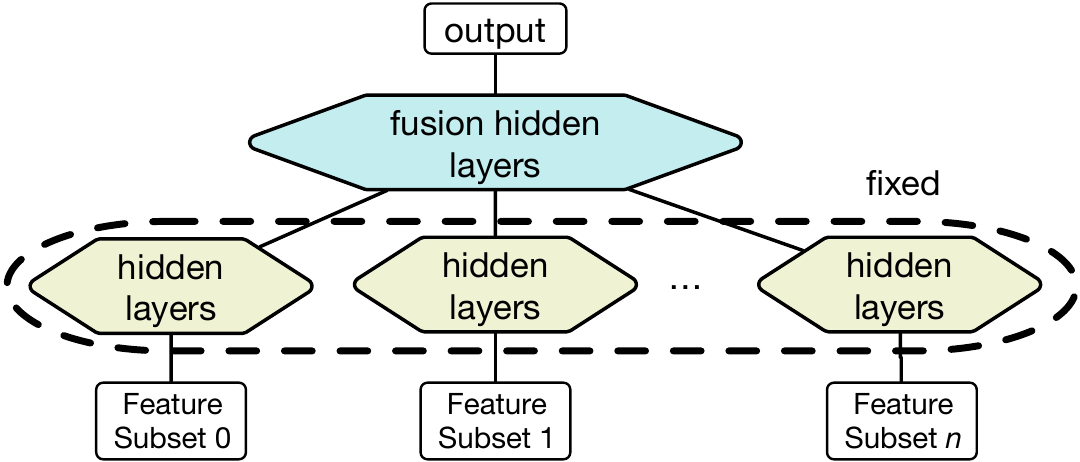}
  \caption{{\it Sparsely-Connected and Disjointly-Trained DNN }}
  \label{fig:SD-DNN}
\end{figure}

To gain both the advantages of small feature sets, which converge to avoid overfitting issues, and to still exploit the redundancy among feature streams, we propose the Sparsely-Connected and Disjointly-Trained DNN (SD-DNN) training framework. 
In this framework, depicted in Fig.~\ref{fig:SD-DNN}, we select a sparse feature set, train (as in the Reduced feature dimensionality DNN's) individual DNN systems. 
Then we fix the parameters of these DNN systems, remove the output layer, connect the top hidden layers together, and add new hidden layers as fusion layers. 
% Haoqi: we connect top hidden layers and add new layers as fusion layers.
This framework allows for both \textit{Sparse Connectivity} at the bottom layers (not all features are connected to all hidden layers above) and \textit{Disjointly Training} the various layers of the DNN thus reducing the degrees of freedom and achieving convergence.

\subsection{Joint Optimization of Sparsely-Connected DNN\label{sec:joint-optim-spars}}
The system presented in the previous section and shown in Fig.~\ref{fig:SD-DNN} disjointly optimizes the sparse lower layers and top fusion layers. 
Without increasing the parameter dimensionality of the SD-DNN, we can initialize training from the disjoint optimization point and jointly optimize the system. We will denote this \textit{Sparse, Jointly} optimized system by SJ-DNN.

\subsection{Local -- Session mappings\label{sec:local-sess-mapp}}
As mentioned earlier, we have only session-level ratings for the couple therapy corpus. This is not unusual in mental health applications given the cost and subjectivity of annotations.

Due to subjectivity and inter-annotator agreement issues we use a binarized subset of the dataset that lies at the top and bottom 20\% of the dataset as in \cite{Black20131} for our training. 
We assign score 1 for high presence and 0 for low presence of one certain behavior. Frame-level training samples are given the same reference as the session level reference as shown on Fig~\ref{fig:OverallProcess}.

At test time, the output of the DNN system provides a score of the presence of behavior (as in Fig.~\ref{fig:DNN_output}), but doesn't provide a global rating. 
While a range of methods exist for fusing decisions (\eg \cite{7178339,xia2015dynamic,lee2012_based-on-isolat}), in this work we will use the simplest one: Average posteriors. 
We can treat the output of DNN, ${q_i^k}$ as a proxy to the posterior probability of the behavior given the frame $i$ for session $k$, and $L_k$ is the number of frames in session $k$. We then average ${q_i^k}$ to derive the session level confidence score ${Q_k}$. Mathematically:
\begin{equation}
{Q_k} = \exp ({1 \over {{L_k}}}\sum\limits_i {\log q_i^k})
\label{scorecal}
\end{equation}
For comparison with the reference session level label, we threshold and binarize ${Q_k}$. The threshold, $T_k$, is selected by optimization to give the minimum classification error rate on the training data. 

\section{Experiment Setup}
We use leave-one-couple-out cross-validation to separate training and test data. We can thus ensure a fair evaluation where same couple is not seen in the test set.
For each behavior code and each gender we use 70 sessions on one extreme of the code (\eg high blame) and 70 sessions at the other extreme (\eg low blame)\footnote{These do not necessarily correspond to matched partners due to the selection of the extreme sessions}. This is to achieve higher inter-annotator agreement and provide training data with binary class labels.

Temporal variation in behavior is slower than basic emotions' and thus a longer frame window size of speech segment is needed for its analysis. 
An earlier work\cite{xia2015dynamic} compared behavior classification performance on various frame sizes and showed that a 20 s frame was sufficient to estimate meaningful behavioral metrics while maintaining high resolution, we thus choose to use a 20 s window with 1 s shift. 

In our experiments we employ 3 of behavioral codes available to us: \textit{Acceptance, Negativity, Blame}. We evaluate using a baseline SVM system and compare with the above proposed DNN based systems.

\emph{In summary:} We use 168 features as discussed in section~\ref{sec:feature}; classify 3 behavioral codes: \textit{Acceptance, Negativity, Blame}; train a 1s-slide, 20s-length rating system; accumulate beliefs towards binary classification evaluation; and qualitatively evaluate the behavioral trajectories resulting from the proposed system.

\section{Results and Discussion}
\Pg{Baseline SVM:} The baseline SVM model was built similar to the Static Behavioral Model discussed in \cite{xia2015dynamic}.\\

\Pg{Fully Connected DNN:} The fully connected DNN system described in section~\ref{sec:dnn-training} did not converge and always kept the first epoch values as the final states. To reduce this issue we had to introduce significant dropout at the input layer. We also had to keep the overall network very small with only one hidden layer of 15 units. We used a mini-batch adaptive gradient optimizer with a mean square error objective function. As seen from Table \ref{final_results}, the fully connected DNN gains were modest.\\

\Pg{Reduced dimensionality DNN:} To create smaller DNNs that may converge easier, we  divided features into 5 parts: (a) knowledge-based split by feature type: pitch, MFCCs, MFBs, jitter and shimmer, intensity. (b) Randomly. 
Then for each feature subset we train a DNN with the same configuration as in the fully connected DNN, \ie one hidden layer with 15 units.

With these reduced and shallow neural nets we immediately observe good training characteristics and convergence. 
Further from the results of Table \ref{tab:category_split} we can observe that even the reduced feature size can often outperform the baseline SVM, which suggests potential gains from employing DNNs for behavior recognition. 
We also note that even the random split can perform quite well in fusion compared to the baseline. Due to the randomness in this feature selection, different splits may even be able to improve, however due to the lack of a development set we decided not to perform such an optimization. 
The knowledge-based feature selection has a less uniform classification accuracy due to the feature-size imbalance as expected, but we obtain better performance on SD-DNN fusion described next, so we use knowledge-based feature split in all following experiments.

\begin{table}[h]
\renewcommand{\arraystretch}{1.5}
\centering
\resizebox{\columnwidth}{!}{%
\renewcommand{\tabcolsep}{0.05cm}
\begin{tabular}{cccccccc}
\toprule
\multicolumn{8}{c}{{ One random feature split instantiation}}\\
\hline
%\parbox{1cm}{ \centering SVM\newline (baseline)} & { subset 0} & { subset 1} & { subset 2} & { subset 3} & { subset 4} & { Fusion} & { SD-DNN }\\ \midrule  
SVM & Subset  & Subset  &  Subset  & Subset  & Subset  &  Fusion &  SD-DNN \\[-1.8ex] 
(Baseline) & 0 & 1 & 2 & 3 & 4 & & \\ \midrule  

%\begin{tabular}[c]{@{}c@{}}SVM\\ (baseline)\end{tabular} & { subset 0} & { subset 1} & { subset 2} & { subset 3} & { subset 4} & { Fusion} & { SD-DNN }\\ \midrule

 { 68.57}  & { 70.36}   & { 72.85}    & { 72.14}    & { 67.50}    & { 67.50}  & { 70.00} &{ 75.00}\\ 
\bottomrule

\toprule
\multicolumn{8}{c}{{ Knowledge-based feature split}}\\
\hline
{ SVM} & { Pitch} & { MFCCs}  & { MFBs}   & { Intensity} & { Jitter \& } & { Fusion} & { SD-DNN}\\[-1.8ex] 
(Baseline) & & & & & Shimmer & & \\ \midrule
{ 68.57}  & { 66.07} & { 71.07} & { 66.78} & { 61.43} & { 61.79} & { 72.14} &{ 75.36} \\ 
\bottomrule
\bottomrule
\end{tabular}}
\caption{Classification accuracy (\%) for the two different feature splits: One random instantiation and one knowledge based}
\label{tab:category_split}
\end{table}

\Pg{SD-DNN:} We thus proceed to construct our SD-DNN system by fixing the parameters of the reduced dimensionality DNN systems and connecting their hidden layers ($15\times5$) to another layer of DNN. In our experiment, we utilize additional two hidden layers with 30 and 10 units respectively, and use the same optimizer and objective function as before. As we can see from the last column of Table\ref{tab:category_split} the performance of the SD-DNN is significantly better than that of the fusion of the individual reduced dimensionality DNN's. 
\\

\Pg{SJ-DNN:} To relax the disjoint optimization constraint we also train jointly reduced feature DNNs at the front layers and the top fusion DNNs of the above model. The parameter space of the model is identical to the SD-DNN except all parameters are initialized on SD-DNN values but jointly trained. Table \ref{final_results} shows that despite the two models being identical, the joint optimization of a larger set of parameters reduces the performance of the SJ-DNN model versus the SD-DNN.\\

\Pg{Fully Connected DNN, SD-DNN Initialized (DNN$_\text{SD-init}$):} After achieving a better performing  system, we attempt once again to reduce sparseness, and hence increase the parameter space of the model, by fully connecting all inputs/hidden layers. We employ the SD-DNN model as initialization instead of using random initialization on DNN. This model is initialized with the weights of the SD-DNN, or zero if the connection did not exist before.
%\panos{table 2 columns should be: SVM, Fully Connected, SD-DNN, SJ-DNN, Fully connected DNN$_\text{SD-init}$}

% discussion & analysis:
All results of experiments are shown in Table \ref{final_results}, in general, the SD-DNN system has higher accuracy rate than SVM baseline and plain DNN system. We obtain the greatest improvement for \textit{Acceptance} behavior from 68.57\% to 75.36\%, which shows benefits in employing DNN and reducing connectivity of DNN because of sparse data.

In summary we can observe that both reduction of the total number of parameters via sparseness but also reduction of the trainable parameters at any time via disjoint training can help in dealing with limited data. 
Specifically by observing the fully connected DNN and DNN$_\text{SD-init}$ results, for most behavioral codes, we can see that any increase in the system's number of parameters (reduction of sparseness) results in reduction of the performance, even if the initialization point is a good one.
We can also see that increasing the number of simultaneously and jointly trainable parameters, as visible by comparing SD- and SJ-DNN's, also damages performance.

%Also, the conclusion that starting with a good initialization point will gain benefits can be verified by comparing 2nd and 5th columns.

%Haoqi: remove positivity-----------------------------
%For the \textit{Positivity} behavior all systems perform poorly, with SVM outperforming all DNN variants. We believe that this is because \textit{Positivity} is not expressed  saliently in the acoustic channel but is rather mostly a function of the semantic content of the interaction.
%\panos{BRIAN QUESTION!!! Haoqi: Pass it on to Brian }
%Nevertheless we do still observe that the sparsely and disjointly trained system outperforms the fully connected DNN.
%Haoqi: remove positivity-----------------------------

\begin{figure}[h!]
  
  \vspace{-0.6cm}
 \centering
   \includegraphics[width=\linewidth]{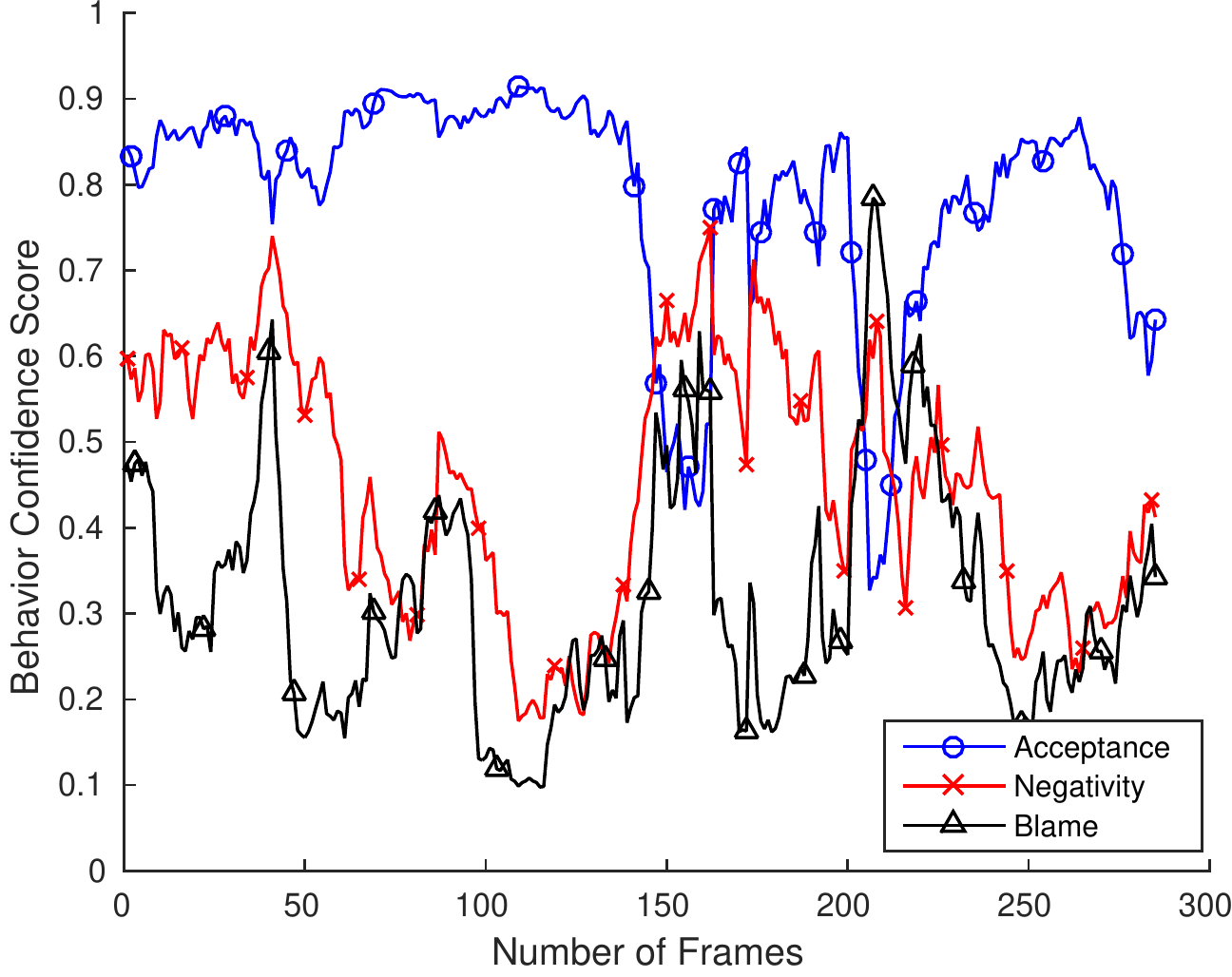}
  %\caption{{\it output of SD-DNN for one certain session with 4 behavior codes }}
  \vspace{-0.8cm}
    \caption{{\it Output of SD-DNN for one sample test session with 3 behavior codes}}
  \label{fig:DNN_output}
\end{figure}

\Pg{Online Behavioral Trajectories:} One of the advantages of moving to an estimation, rather than classification framework, is that we can now provide domain experts with behavioral trajectories. 
These are becoming increasingly necessary, especially in new behavioral analysis paradigms where patients are instrumented continuously in-lab, at-home, and \textit{in-situ}. 
The resulting datasets are vast, even though training data is limited, and behavioral trajectories can help identify specific behaviors over time. 
One sample behavior dynamic change trajectories is shown in Fig.~\ref{fig:DNN_output}. 
From this figure, we can see behavior \textit{Negativity} and \textit{Blame} are highly correlated, and have opposite trend with \textit{Acceptance}, which is in agreement with our intuition and previous research work\cite{Black20131}. 
%Also, we can find that acoustic features cannot reflect \textit{Positivity} change, which leads its lower classification accuracy.

Overall, results suggest that a Sparsely-Connected, Disjointly-Trained DNN framework provides the most promise in employing DNNs into the limited data BSP domain.

\begin{table}[h!]
\renewcommand{\arraystretch}{0.9}
\centering
\resizebox{\columnwidth}{!}{%
\renewcommand{\tabcolsep}{0.12cm}
\begin{tabular}{cccccc}
\toprule
%\begin{tabular}[c]{@{}c@{}}behavior\\ code\end{tabular} & SVM & \begin{tabular}[c]{@{}c@{}}Fully connected\\ DNN\end{tabular} & \begin{tabular}[c]{@{}c@{}}DNN AAA\\  connected\end{tabular} & SJ-DNN & SD-DNN \\ \midrule

\begin{tabular}[c]{@{}c@{}}Behavior\\ Code\end{tabular} & SVM & \begin{tabular}[c]{@{}c@{}}Fully\\ connected\\ DNN\end{tabular} & SD-DNN & SJ-DNN & DNN$_\text{SD-init}$ \\[1.5ex] \midrule
Acceptance & 68.57 & 71.79 & \textbf{75.36} & 73.57 & 71.43 \\[1.5ex]
Negativity & 73.21 & 74.64 & \textbf{77.14} & 75.36 & 74.29 \\[1.5ex]
%Positivity & \textbf{65.71} & 63.57 & 63.90 & 61.43 & 63.21 \\
Blame      & 73.21 & 73.93 & \textbf{75.71} & 74.29 & 73.93 \\
%Acceptance & 68.57 & 71.79 & 71.43 & 73.57 & 75.36 \\
%Negativity & 73.21 & 74.64 & 74.29 & 75.36 & 77.14 \\
%Positivity & 65.71 & 63.57 & 63.21 & 61.43 & 63.90 \\ 
%Blame      & 73.21 & 73.93 & 73.93 & 74.29 & 75.71 \\ 
\bottomrule
\end{tabular}
}
\caption{Classification accuracy (\%) with all behavioral recognition systems}
\label{final_results}
\end{table}
 
\section{Conclusion and Future Work}    
Compared to other DNN based machine learning tasks, data sparsity is a critical issue in BSP domain due to its costly and complicated data generating process. Through \textit{Sparsely Connected} and \textit{Disjoint Training} we can train more complex architecture DNN systems with limited dataset, achieve increased session-level performance, and importantly obtain continuous in time and rating annotations of our data. 

For future work, we plan to employ mutual or shared information between different behavior codes into behavioral analysis, since some behaviors are highly correlated. Also, we will tune the SD-DNN architecture and parameters. For instance, different reduced dimensionality DNN learning system can use different DNN architecture.
 
  \newpage
  \eightpt
  \bibliographystyle{IEEEtran}

  \bibliography{mybib}

% Generated by IEEEtran.bst, version: 1.13 (2008/09/30)
\begin{thebibliography}{10}
\providecommand{\url}[1]{#1}
\csname url@samestyle\endcsname
\providecommand{\newblock}{\relax}
\providecommand{\bibinfo}[2]{#2}
\providecommand{\BIBentrySTDinterwordspacing}{\spaceskip=0pt\relax}
\providecommand{\BIBentryALTinterwordstretchfactor}{4}
\providecommand{\BIBentryALTinterwordspacing}{\spaceskip=\fontdimen2\font plus
\BIBentryALTinterwordstretchfactor\fontdimen3\font minus
  \fontdimen4\font\relax}
\providecommand{\BIBforeignlanguage}[2]{{%
\expandafter\ifx\csname l@#1\endcsname\relax
\typeout{** WARNING: IEEEtran.bst: No hyphenation pattern has been}%
\typeout{** loaded for the language `#1'. Using the pattern for}%
\typeout{** the default language instead.}%
\else
\language=\csname l@#1\endcsname
\fi
#2}}
\providecommand{\BIBdecl}{\relax}
\BIBdecl

\bibitem{el2011survey}
M.~El~Ayadi, M.~S. Kamel, and F.~Karray, ``Survey on speech emotion
  recognition: Features, classification schemes, and databases,'' \emph{Pattern
  Recognition}, vol.~44, no.~3, pp. 572--587, 2011.

\bibitem{Vogt2008}
T.~Vogt, E.~Andr{\'e}, and J.~Wagner, \emph{Affect and Emotion in
  Human-Computer Interaction: From Theory to Applications}.\hskip 1em plus
  0.5em minus 0.4em\relax Berlin, Heidelberg: Springer Berlin Heidelberg, 2008,
  ch. Automatic Recognition of Emotions from Speech: A Review of the Literature
  and Recommendations for Practical Realisation, pp. 75--91.

\bibitem{Schuller20111062}
B.~Schuller, A.~Batliner, S.~Steidl, and D.~Seppi, ``Recognising realistic
  emotions and affect in speech: State of the art and lessons learnt from the
  first challenge,'' \emph{Speech Communication}, vol.~53, no. 9{\^a}€``10,
  pp. 1062 -- 1087, 2011, sensing Emotion and Affect - Facing Realism in Speech
  Processing.

\bibitem{6296526}
G.~Hinton, L.~Deng, D.~Yu, G.~E. Dahl, A.~r.~Mohamed, N.~Jaitly, A.~Senior,
  V.~Vanhoucke, P.~Nguyen, T.~N. Sainath, and B.~Kingsbury, ``Deep neural
  networks for acoustic modeling in speech recognition: The shared views of
  four research groups,'' \emph{IEEE Signal Processing Magazine}, vol.~29,
  no.~6, pp. 82--97, Nov 2012.

\bibitem{5947651}
A.~Stuhlsatz, C.~Meyer, F.~Eyben, T.~Zielke, G.~Meier, and B.~Schuller, ``Deep
  neural networks for acoustic emotion recognition: Raising the benchmarks,''
  in \emph{Acoustics, Speech and Signal Processing (ICASSP), 2011 IEEE
  International Conference on}, {May} 2011, pp. 5688--5691.

\bibitem{Schuller2015}
B.~Schuller, \emph{Advances in Neural Networks: Computational and Theoretical
  Issues}.\hskip 1em plus 0.5em minus 0.4em\relax Cham: Springer International
  Publishing, 2015, ch. Deep Learning Our Everyday Emotions, pp. 339--346.

\bibitem{han2014speech}
K.~Han, D.~Yu, and I.~Tashev, ``Speech emotion recognition using deep neural
  network and extreme learning machine.'' in \emph{Interspeech}, 2014, pp.
  223--227.

\bibitem{6707732}
D.~Le and E.~M. Provost, ``Emotion recognition from spontaneous speech using
  hidden markov models with deep belief networks,'' in \emph{Automatic Speech
  Recognition and Understanding (ASRU), 2013 IEEE Workshop on}, {Dec} 2013, pp.
  216--221.

\bibitem{6457406}
S.~Narayanan and P.~G. Georgiou, ``Behavioral signal processing: Deriving human
  behavioral informatics from speech and language,'' \emph{Proceedings of the
  IEEE}, vol. 101, no.~5, pp. 1203--1233, {May} 2013.

\bibitem{Georgiou:2011:BSP}
P.~G. Georgiou, M.~P. Black, and S.~S. Narayanan, ``Behavioral signal
  processing for understanding (distressed) dyadic interactions: Some recent
  developments,'' in \emph{Proceedings of the 2011 Joint ACM Workshop on Human
  Gesture and Behavior Understanding}, ser. J-HGBU '11.\hskip 1em plus 0.5em
  minus 0.4em\relax New York, NY, USA: ACM, 2011, pp. 7--12.

\bibitem{Georgiou2011}
P.~G. Georgiou, M.~P. Black, A.~C. Lammert, B.~R. Baucom, and S.~S. Narayanan,
  \emph{Affective Computing and Intelligent Interaction: 4th International
  Conference, ACII 2011, Memphis, TN, USA, October 9--12, 2011, Proceedings,
  Part I}.\hskip 1em plus 0.5em minus 0.4em\relax Berlin, Heidelberg: Springer
  Berlin Heidelberg, 2011, ch. ``That's Aggravating, Very Aggravating'': Is It
  Possible to Classify Behaviors in Couple Interactions Using Automatically
  Derived Lexical Features?, pp. 87--96.

\bibitem{Black20131}
M.~P. Black, A.~Katsamanis, B.~R. Baucom, C.-C. Lee, A.~C. Lammert,
  A.~Christensen, P.~G. Georgiou, and S.~S. Narayanan, ``Toward automating a
  human behavioral coding system for married couples' interactions using speech
  acoustic features,'' \emph{Speech Communication}, vol.~55, no.~1, pp. 1 --
  21, 2013.

\bibitem{7106538}
B.~Xiao, P.~Georgiou, B.~Baucom, and S.~S. Narayanan, ``Head motion modeling
  for human behavior analysis in dyadic interaction,'' \emph{IEEE Transactions
  on Multimedia}, vol.~17, no.~7, pp. 1107--1119, July 2015.

\bibitem{6607640}
A.~Metallinou, R.~B. Grossman, and S.~Narayanan, ``Quantifying atypicality in
  affective facial expressions of children with autism spectrum disorders,'' in
  \emph{Multimedia and Expo (ICME), 2013 IEEE International Conference on},
  {July} 2013, pp. 1--6.

\bibitem{7178339}
S.~N. Chakravarthula, R.~Gupta, B.~Baucom, and P.~Georgiou, ``A language-based
  generative model framework for behavioral analysis of couples' therapy,'' in
  \emph{Acoustics, Speech and Signal Processing (ICASSP), 2015 IEEE
  International Conference on}, {April} 2015, pp. 2090--2094.

\bibitem{xia2015dynamic}
W.~Xia, J.~Gibson, B.~Xiao, B.~Baucom, and P.~G. Georgiou, ``A dynamic model
  for behavioral analysis of couple interactions using acoustic features,'' in
  \emph{Sixteenth Annual Conference of the International Speech Communication
  Association}, 2015.

\bibitem{opensmile}
F.~Eyben, F.~Weninger, F.~Gross, and B.~Schuller, ``Recent developments in
  opensmile, the munich open-source multimedia feature extractor,'' in
  \emph{Proceedings of the 21st ACM International Conference on Multimedia},
  ser. MM '13.\hskip 1em plus 0.5em minus 0.4em\relax New York, NY, USA: ACM,
  2013, pp. 835--838.

\bibitem{boersma2002praat}
P.~Boersma \emph{et~al.}, ``Praat, a system for doing phonetics by computer,''
  \emph{Glot international}, vol.~5, no. 9/10, pp. 341--345, 2002.

\bibitem{christensen2004traditional}
A.~Christensen, D.~C. Atkins, S.~Berns, J.~Wheeler, D.~H. Baucom, and L.~E.
  Simpson, ``Traditional versus integrative behavioral couple therapy for
  significantly and chronically distressed married couples.'' \emph{Journal of
  consulting and clinical psychology}, vol.~72, no.~2, p. 176, 2004.

\bibitem{heavey2002couples}
C.~Heavey, D.~Gill, and A.~Christensen, ``Couples interaction rating system 2
  (cirs2),'' \emph{University of California, Los Angeles}, vol.~7, 2002.

\bibitem{jones1998couples}
J.~Jones and A.~Christensen, ``Couples interaction study: Social support
  interaction rating system,'' \emph{University of California, Los Angeles},
  vol.~7, 1998.

\bibitem{dropout}
N.~Srivastava, G.~Hinton, A.~Krizhevsky, I.~Sutskever, and R.~Salakhutdinov,
  ``Dropout: A simple way to prevent neural networks from overfitting,''
  \emph{J. Mach. Learn. Res.}, vol.~15, no.~1, pp. 1929--1958, Jan. 2014.

\bibitem{lee2012_based-on-isolat}
C.-C. Lee, A.~Katsamanis, P.~G. Georgiou, and S.~S. Narayanan, ``Based on
  isolated saliency or causal integration? toward a better understanding of
  human annotation process using multiple instance learning and sequential
  probability ratio test,'' in \emph{Proceedings of InterSpeech}, Sep. 2012.

\end{thebibliography}

\end{document}